%% file: main.tex
\documentclass[letterpaper]{article} 
\usepackage{aaai2026}  
\usepackage{times}  
\usepackage{helvet}  
\usepackage{courier}  
\usepackage[hyphens]{url}  
\usepackage{graphicx} 
\usepackage{natbib}  
\usepackage{caption} 
\usepackage{algorithm}
\usepackage{algorithmic}

\usepackage{newfloat}
\usepackage{listings}
\DeclareCaptionStyle{ruled}{labelfont=normalfont,labelsep=colon,strut=off} 
\floatstyle{ruled}
\newfloat{listing}{tb}{lst}{}
\floatname{listing}{Listing}
\usepackage{graphicx}

\usepackage{amsmath}
\usepackage{amssymb}
\usepackage{mathtools}
\usepackage{amsthm}
\usepackage{mathbbol}
\usepackage{multirow}
\usepackage{graphicx}
\usepackage{subfigure}
\usepackage{booktabs} 
\usepackage{subfigure}
\usepackage{booktabs} 
\usepackage{dsfont}
\usepackage{mathtools}
\usepackage{amsfonts}
\usepackage{xcolor}
\usepackage{varwidth}
\usepackage{amsmath}
\usepackage{amsthm}
\usepackage{mathtools,amssymb}
\usepackage{lscape}
\usepackage{enumitem}
\usepackage{stmaryrd}

\newtheorem{theorem}{Theorem}

\newtheorem{lemma}{Lemma}

\newtheorem{definition}{Definition}
\newtheorem{corollary}{Corollary}

\newtheorem{lemma_appendix}{Lemma}[section]
\newtheorem{theorem_appendix}{Theorem}[section]
\newtheorem{corollary_appendix}{Corollary}[section]

\def\P{\mathbb P}
\def\R{\mathbb R}

\def\st{\text{s.t.}}

\def\indicator{\mathbb{1}}

\def\calX{\mathcal X}
\def\calY{\mathcal Y}

\def\calP{\mathcal P}

\def\calU{\mathcal U}

\def\P{\mathbb P}
\def\R{\mathbb R}

\def\min{\text{min}}

\def\max{\text{max}}

\def\cal{\text{cal}}
\def\test{\text{test}}

\title{ Provably Minimum-Length Conformal Prediction Sets for Ordinal Classification }
\author{
    Zijian Zhang\equalcontrib\textsuperscript{\rm 1},
    Xinyu Chen\equalcontrib\textsuperscript{\rm 1},
    Yuanjie Shi\textsuperscript{\rm 1},
    Liyuan Lillian Ma\textsuperscript{\rm 2},
    Zifan Xu\textsuperscript{\rm 3},
    Yan Yan\thanks{Corresponding author}\textsuperscript{\rm 1}
}
\affiliations{
    \textsuperscript{\rm 1}School of Electrical Engineering and Computer Science, Washington State University\\
    \textsuperscript{\rm 2}Independent researcher
    \textsuperscript{\rm 3}Independent researcher

    \{zijian.zhang2, xinyu.chen1, yuanjie.shi, yan.yan1\}@wsu.edu, liyuanma2015@gmail.com, zfan0780@gmail.com
}

\usepackage{bibentry}

\begin{document}

\maketitle

\begin{abstract}
Ordinal classification has been widely applied in many high-stakes applications, e.g., medical imaging and diagnosis,
where reliable uncertainty quantification (UQ) is essential for decision making.
Conformal prediction (CP) is a general UQ framework that provides statistically valid guarantees,
which is especially useful in practice. 
However, prior ordinal CP methods mainly focus on heuristic algorithms or restrictively require the underlying model to predict a unimodal distribution over ordinal labels. 
Consequently, they provide limited insight into {\it coverage–efficiency trade-offs}, 
or a {\it model-agnostic and distribution-free} nature favored by CP methods.
To this end, we fill this gap by propose an ordinal-CP method that is model-agnostic and provides {\it instance-level optimal} prediction intervals.
Specifically, we formulate conformal ordinal classification as a minimum-length covering problem at the instance level.
To solve this problem, we develop a sliding-window algorithm that is optimal on each calibration data,
with only a linear time complexity in $K$, the $\#$ of label candidates.
The local optimality per instance further also improves predictive efficiency in expectation.
Moreover, we propose a length-regularized variant that shrinks prediction set size while preserving coverage.
Experiments on three benchmark datasets from diverse domains are conducted to demonstrate the significantly improved predictive efficiency of the proposed methods over baselines (by 15\%{$\downarrow$} on average over three datasets).
\end{abstract}

\textbf{Code - }https://github.com/xrty/OCP


\section{Introduction}
\label{section:introduction}

Ordinal classification plays a significant role in a wide range of real-world applications, e.g., medical diagnosis \cite{albuquerque2021ordinal}, credit risk assessment \cite{kwon1997ordinal,fernandez2013addressing,hirk2019multivariate}, and age estimation \cite{niu2016ordinal}. 
Unlike standard classification tasks that treat labels as unordered categories, 
ordinal classification explicitly models the inherent order among discrete labels. 
On one hand, this ordinal structure allows models to produce more interpretable predictions by reflecting the natural ranking in the output \cite{gutierrez2015ordinal}.
On the other hand, when the labels from the underlying distribution exhibit a clear ordinal structure,
ordinal classification allows us to leverage the ordering information over classes, leading to better theoretical guarantees and empirical performance, e.g., Fisher consistency \cite{pedregosa2017consistency}.

While ordinal classification is highly valuable in real-world tasks, its practical effectiveness is often limited by its ability of uncertainty quantification (UQ) \cite{cresswell2024conformal,straitouri2023improving}. 
In the absence of uncertainty-aware mechanisms, ordinal models typically produce a single point prediction without indicating statistically calibrated confidence. 
This can be problematic in safety-critical applications, especially when making a decision.
For instance, in medical imaging, a model might assign a patient to a specific disease severity level based on subtle features \cite{zhang2024new}.
However, if the model is poorly calibrated or highly uncertain due to limited data or domain shifts, such a point prediction may be misleading \cite{guo2017calibration}. 
Without a principled way to express uncertainty, practitioners may either over-rely on the model or disregard its output altogether, undermining its role in decision support. 
Therefore, integrating UQ into ordinal classification is essential to ensure reliable and interpretable deployment in high-stakes scenarios.

Conformal Prediction (CP) \cite{vovk1999machine,vovk2005algorithmic,shafer2008tutorial,angelopoulos2021gentle} is a post-hoc framework that provides distribution-free prediction sets with guaranteed marginal coverage (aka. statistically valid coverage), regardless of the underlying model's confidence. 
Instead of committing to a single label as prediction, CP constructs a set of plausible labels for each instance, offering a principled way to quantify uncertainty. 
It is especially valuable in ordinal classification, 
where prediction confidence may vary significantly across classes 
and misclassifications may carry asymmetric consequences.

However, despite several efforts, extending standard CP to ordinal classification remains highly non-trivial.
For example, \cite{lu2022ordinalaps} adapts the seminal adaptive prediction sets (APS) \cite{romano2020classification},
which is originally designed for standard classification, 
and proposes ordinal APS.
Their method differs from APS in the way to construct the prediction set -- 
given an input, it starts from the most probable label, 
and iteratively expands the prediction set by including the more confident neighboring label (either to the left or right),
until the cumulative confidence exceeds a threshold calibrated on held-out calibration data.
Nevertheless, this greedy search algorithm is inherently heuristic: 
it does not provide optimality guarantees and can yield larger-than-necessary intervals in some cases (i.e., worse predictive efficiency).

Another line of work, COPOC \cite{dey2023ordinalcp}, enforces unimodality on the model's predicted distribution by learning an auxiliary transformation sub-module.
While the ordinal structure encourages contiguous prediction sets, 
it still relies on a unimodality-enforcing module, which reduces the method’s model-agnostic nature. 

To address the two aforementioned challenges for ordinal conformal prediction, we propose a novel algorithm, minimum-length conformal prediction sets (min-CPS).
Analogous to prior methods, min-CPS aims to minimize the expected prediction interval subject to the marginal coverage constraint.
However, instead of relying on heuristic or assuming model-specific conditions, 
min-CPS introduces a computationally efficient sliding-window algorithm that guarantees an exact solution to the minimum-length prediction set for each individual instance.
In addition, we propose a length-regularized variant, i.e., min-RCPS, which explicitly incorporates the ordinal structure to further enhance efficiency, 
especially under uncertain intervals. 
This variant trades off between coverage tightness and structural alignment, leading to more semantically coherent prediction sets.

The main contributions of this paper are summarized as:
\begin{itemize}
\item We propose a novel algorithm, minimum-length conformal prediction sets (min-CPS), 
along with a length-regularized variant (min-RCPS) that leverages ordinal structure to improve predictive efficiency by balancing coverage tightness and semantic coherence.

\item We provide theoretical guarantees for min-CPS, proving its instance-level optimality of the constructed intervals and establishing conditions that ensure monotonic empirical coverage.
We empirically verify the monotonicity condition of the empirical coverage, under which a valid coverage is guaranteed.


\item We conduct comprehensive experiments on three benchmark datasets, demonstrating that min-CPS consistently outperforms existing conformal prediction methods for ordinal classification in both predictive efficiency and interval coherence (min-CPS $14\%${$\downarrow$} and min-RCPS $15\%${$\downarrow$} reduction in the average length of prediction intervals).
\end{itemize}

\section{Related Work}
\label{sec:related}

\paragraph{Ordinal Classification,}
also known as ordinal regression, addresses problems where the labels have a natural ordering but unknown spacing between categories \cite{WangSurvey2025,gutierrez2015ordinal}. 
Unlike traditional multiclass classification, ordinal methods aim to respect this inherent order to improve predictive performance and interpretability. 
It has been widely used in real-world applications,
e.g., facial age estimation \cite{Shin2022,wang2023ord2seq},
aesthetics assessment \cite{he2022rethinking,kong2016photo}, 
diabetic retinal grading \cite{yu2024clip,cheng2023robust} and monocular depth estimation \cite{shao2023iebins}.

\paragraph{Conformal Prediction (CP)}
equips any base learner with \emph{finite-sample, distribution-free} guarantees on predictive uncertainty, and has drawn increasing attention recently.
Classical “full’’ CP computes leave-one-out non-conformity scores and inverts a series of hypothesis tests to ensure marginal coverage \(1-\alpha\) on unseen data \cite{vovk2005algorithmic,shafer2008tutorial}.
Some studies focus on cross-validation methods \cite{vovk2015cross} and jackknife+ approach \cite{barber2021predictive},
while its split (inductive) variant attains near-linear inference by calibrating a single quantile on a held-out set \cite{lei2018distribution,romano2020classification,oliveira2024split},
in which calibration is completed on a held-out dataset.
Building on this foundation, researchers have devised score functions that tighten prediction sets without sacrificing validity.
These examples include
Adaptive Prediction Sets (APS) \cite{romano2020classification}, regularized APS \cite{angelopoulos2020uncertainty},  
SAPS \cite{huang2024labelrank}, 
Clustered CP \cite{ding2024class},
PoT-CP \cite{huang2025conformal},
RC3P \cite{shi2024conformal}, etc.
For regression tasks, Conformalized Quantile Regression delivers valid confidence intervals \cite{romano2019conformalized}.

Recent work further refines CP for modern deep-learning by information-theoretic objectives \cite{correia2024information}, 
self-calibrating Venn–Abers schemes \cite{vanderlaan2024selfcal},  probabilistic relaxations toward conditional validity \cite{plassier2024probabilistic},
robustness \cite{gendler2021adversarially,ghosh2023probabilistically,liu2024pitfalls},
and generic conditional validity \cite{feldman2021improving,gibbs2023conformal,vovk2012conditional,kiyani2024length}.
Domain-specific extensions now support large-language-model inference with or without logit access \cite{cherian2024llmvalidity,su2024api},
graph neural networks \cite{zhang2025residual},
vision model \cite{chen2025conformalsam},
and enhance driver perception in adverse conditions through augmented-reality CP \cite{doula2024arcp}. 
A recent trend also arises to integrate CP into training \cite{stutz2022learning,shi2025direct} to improve the model uncertainty.
Together, these advances demonstrate CP’s versatility and its central role in uncertainty-aware machine learning.

\paragraph{Ordinal CP.}
Standard CP methods do not exploit the ordered structure of labels, often yielding
non-contiguous prediction sets that are ill-suited to ordinal tasks.
\emph{Ordinal Adaptive Prediction Sets} (Ordinal-APS) introduce a score
that accumulates \emph{contiguous} soft-max probabilities, guaranteeing
coverage while respecting the class order \cite{lu2022ordinalaps}.  Building on this idea,
\cite{dey2023ordinalcp} cast the problem under a unimodality assumption and
prove tight bounds on the minimal contiguous set size.
Beyond coverage, \cite{xu2023crcoc}
control more general risk measures (e.g., expected loss) in the ordinal setting,
while \cite{chakraborty2024dfocp} leverage multiple-testing theory to construct
both contiguous and non-contiguous sets with family-wise error control.

\section{Proposed Min-CPS and Min-RCPS}
\label{section:method}

In this section, we introduce our proposed algorithms, minimum-length conformal prediction sets (min-CPS) and its length-regularized variant minimum-length regularized conformal prediction sets (min-RCPS).
We begin by the key notations and definitions used throughout the paper in \textsection{\ref{subsection:notations}}. 
Next, in \textsection{\ref{subsection:instance_level_min_length_covering}}, we formulate the core building block, minimum-length covering formulation in the instance level,
which serves as the foundation for both methods.
We then show in \textsection\ref{subsection:min_CPS} how this local formulation leads to global optimality of min-CPS in the population level, yielding provably efficient conformal prediction sets.
Finally, in \textsection\ref{subsection:min_RCPS}, we describe the regularized variant min-RCPS.

\subsection{ Notations and Problem Setup }
\label{subsection:notations}

\paragraph{Notations.}
Suppose $X \in \calX$ is an input from the input space $\calX$, and $Y \in \calY = \{ 1,2,\cdots, K\}$ is the ordinal ground-truth label, where $K$ is the number of candidate classes. 
Assume that $(X, Y)$ is a data sample drawn from an underlying distribution $\calP$ defined on the joint space $\calX \times \calY$. 
Let $f(X): \calX \rightarrow \Delta_+^K$ denote an underlying pretrained model that predicts confidence for each class label,
where $\Delta_+^K$ is the (K-1)-dimensional probability simplex,
$f(X)_y$ represents the predicted confidence score of class $y \in \calY$.
Define $D_\cal = \{ (X_i, Y_i) \}_{i=1}^n$ as the calibration set of $n$ examples,
while $D_\test = \{ (X_i, Y_i) \}_{i=n+1}^{n+m}$ denotes the test set of $m$ samples, all being exchangeable.
Define $\indicator[\cdot]$ as an indicator function.
$\mathbb N$ represents the set of natural numbers.

\paragraph{Conformal Prediction for Ordinal Classification.}
Analogous to standard CP methods, ordinal CP algorithms aim to guarantee the following marginal coverage:
\begin{align}\label{eq:marginal_coverage_in_ordinal_classification}
&
\P_{XY} \{ Y \in \widehat C_\tau(X) \}
\geq
1 - \alpha
, 
\\
& \quad  ~~~
\text{ where }
\widehat C_\tau(X) = \{ y \in \calY : l(X; \tau) \leq y \leq u(X; \tau) \}
,
\nonumber
\end{align}
where $l, u: \calX \times \R \rightarrow \calY$ are some functions that map the input $X$ and a calibration threshold $\tau$ to lower and upper bounds of the prediction interval, respectively.

In the context of ordinal classification, it is natural to impose the additional structural constraint that the prediction set forms a continuous interval over the label space. 
This reflects the semantic ordering of the labels and enhances interpretability, especially in high-stakes applications such as risk stratification or severity grading.
If we relax the structure and allow $\widehat C: \calX \rightarrow \{0, 1\}^{|\calY|}$ to include any arbitrary subset of $\calY$ (not necessarily a continuous interval), 
then the framework reduces to a standard CP for multiclass classification, 
where the label order is ignored.

One notable exception arises when the model's predicted probability distribution is unimodal over $\calY$.
In this case, the standard CP methods naturally produce a contiguous prediction set on any $X$, 
since they sequentially include class labels in descending order of confidence, 
resulting in a single and connected interval around the mode (the most confident label).
This property serves as the key motivation behind \cite{dey2023ordinalcp},
which explicitly enforces unimodality on predicted probability, 
so that standard CP methods can be directly applied.

Typically, the lower and upper functions $l$ and $u$ satisfy the {\it nestedness} property \cite{vovk2005algorithmic,shafer2008tutorial} s.t. (with nonconformity scores):
\begin{align*}
\tau_1 \leq \tau_2 
\Rightarrow
\widehat C_{\tau_2}(X) \subseteq \widehat C_{\tau_1}(X)
,
\end{align*}
which means that smaller thresholds lead to wider (more conservative) prediction intervals. 
This monotonicity property is a standard requirement in CP \cite{vovk2005algorithmic,shafer2008tutorial}, 
and is essential for calibrating the prediction set size to achieve the desired marginal coverage.

Accordingly, prior ordinal CP methods \cite{lu2022ordinalaps,dey2023ordinalcp} share the same objective:
to determine the smallest (least conservative) threshold $\tau$ on the held-out calibration set $D_\cal$ that guarantees the desired coverage,
thus formulated as:
\begin{align}\label{eq:ordinal_cp_without_considering_predictive_efficiency}
\min & ~~ \tau 
\\
\st &
\sum_{i=1}^n \indicator [ l(X_i; \tau) \leq Y_i \leq u(X_i; \tau) ] \geq (1 - \alpha) (n+1)
,
\nonumber
\end{align}
which can be interpreted as {\it finding the least conservative coverage (by maximizing $\tau$) that guarantee $1-\alpha$ coverage}.
Under the condition of data exchangeability, any $\tau$ satisfying the above constraint yields a conformal predictor with valid marginal coverage at level $1 - \alpha$ as in (\ref{eq:marginal_coverage_in_ordinal_classification}).

\subsection{ Instance-Level Minimum-Length Covering }
\label{subsection:instance_level_min_length_covering}

Although solving (\ref{eq:ordinal_cp_without_considering_predictive_efficiency}) ensures the valid marginal coverage,
existing studies have not provided insight into predictive efficiency (e.g., the expected size of the prediction intervals).
For example, \cite{lu2022ordinalaps} proposes a heuristic procedure for searching the lower and upper bounds of the prediction interval. 
However, the optimality gap of this approach remains under-explored, 
and it is unclear how close the resulting prediction sets are to the minimum possible length (best possible predictive efficiency).
Hence, {\it it is still unclear whether existing ordinal CP methods reach the optimal predictive efficiency, 
and if not, what algorithms can be used to attain the optimum}.

\paragraph{Minimum-Length Covering on Each Instance.}
For any input $X \in \calX$ and threshold $\tau \in (0,1)$, 
we cast the search for the shortest interval as the following minimum-length covering problem:
\begin{align}\label{eq:instance_coverage_problem}
&
\min_{(l, u) \in \calU(X; \tau)} ~ 
\underbrace{ \ell(l, u) \triangleq u - l
}_{ \text{interval length} },
~
\text{ \st }~
\tau \in (0, 1), \text{ and}
\\
&
\calU(X; \tau) 
= 
\Bigg\{
(l, u)
: 
\underbrace{
\sum_{k=l}^u f(X)_k 
}_{ \text{(i) covering prob.} }
\geq \tau,
\underbrace{
l \leq \hat y^*(X) \leq u 
}_{ \text{(ii) including anchor} }
\Bigg\}
,
\nonumber
\end{align}
where $\hat y^*(X)$ is the mode of the predicted distribution (i.e., the most confident predicted label), 
and $\calU(X; \tau)$ denotes a feasible set which requires the interval $[l, u]$ to (i) cover sufficient probability mass at least $\tau$,
and (ii) include the anchor label $\hat y^*(X)$.

\begin{algorithm}[t]
\caption{ Instance-Level Minimum-Length Covering }
\label{alg:instance_level_min_length_covering}
\begin{algorithmic}[1]
\REQUIRE Probabilities $\{f(X)_k\}_{k=1}^K$ on $X$, 
a threshold $\tau \in (0, 1)$, 
the mode $\hat y^*(X) = \arg\max_{k \in \calY} f(X)_k$

\STATE Initialization: initial lower bound $l \gets 1$, prefix sum $P_0 \gets 0$, output bounds $(l^*, u^*) \gets (-1, -1)$, output length $\ell^* \gets \infty$

\FOR{$k=1, ..., K$}
\label{alg:instance_level_min_length_covering:line:prefix_sum_start}

  \STATE $P_k \gets P_{k-1} + f(X)_k$  \hfill // Prefix sum array

\ENDFOR
\label{alg:instance_level_min_length_covering:line:prefix_sum_end}

\FOR{$u = \hat y^*(X), ..., K$}
\label{alg:instance_level_min_length_covering:line:outer_loop}

    \WHILE{$l \leq \hat y^*(X)$ \AND $\underbrace{P_u - P_{l-1}}_{ \text{prob. within }[l,u] } \geq \tau$}
    \label{alg:instance_level_min_length_covering:line:inner_loop}
    
        \IF{$u - l < \ell^*$ } 
            \STATE $(l^*, u^*) \gets (l, u)$, $\ell^*\gets u - l$ 
        \ENDIF
        \STATE $l \gets l + 1$
    \ENDWHILE
\ENDFOR
\RETURN Output bounds $(l^*, u^*)$ for prediction interval
\end{algorithmic}
\end{algorithm}

\paragraph{Efficient Sliding-Window Algorithm.}
A na\"ive brute-force search examines all ${K\choose 2}+K=O(K^{2})$ possible candidate pairs of $(l, u)$,
where the ${K \choose 2}$ possible pairs come from the case that $l \neq u$,
while the additional $K$ possible pairs are from $l = u$.
Instead, we design a linear-time sliding-window algorithm (summarized in Algorithm \ref{alg:instance_level_min_length_covering}) that enumerates only a minimal sequence of feasible intervals satisfying the two constraints,
i.e., covering sufficient probability and including the anchor label.

\vspace{0.5ex}
\noindent
{\bf (i) The constraint of covering sufficient probability.}
We pre-compute prefix sums $P_k = \sum_{i=1}^k f(X)_i$ (Line \ref{alg:instance_level_min_length_covering:line:prefix_sum_start} - 
\ref{alg:instance_level_min_length_covering:line:prefix_sum_end}in Algorithm \ref{alg:instance_level_min_length_covering}),
so that the probability mass of any interval $[l, u]$ is computed by $P_u - P_{l-1}$ (Line \ref{alg:instance_level_min_length_covering:line:inner_loop}) in $O(1)$ time complexity.
Starting from $u=\hat y^*(X)$, we increase $u$ monotonically to $K$  (Line \ref{alg:instance_level_min_length_covering:line:outer_loop}),
and advance $l$ only while coverage is reached (Line \ref{alg:instance_level_min_length_covering:line:inner_loop}),
which ensures each boundary moves at most once throughout Algorithm \ref{alg:instance_level_min_length_covering}.

\vspace{0.5ex}
\noindent
{\bf (ii) The constraint of including anchor.}
We restrict attention to intervals containing the anchor label by enforcing
$l\le\hat y^*(X)\le u$ implicitly:
$u$ is initialized at $\hat y^*(X)$
(Line~\ref{alg:instance_level_min_length_covering:line:outer_loop})
and the inner loop is guarded by $l\le\hat y^*(X)$
(Line~\ref{alg:instance_level_min_length_covering:line:inner_loop}).

These two design choices reduce the search to
$O(K)$ range-sum evaluations, yielding the exact optimum in linear time.
Formally, we present the following theorem to summarize our theoretical result.
\begin{theorem}
\label{theorem:complexity_minimum_length_instance_coverage}
(Optimality and complexity of Algorithm \ref{alg:instance_level_min_length_covering})
Let $K \in \mathbb N$ and $\tau \in (0, 1]$.
For any input $X \in \calX$,
Algorithm \ref{alg:instance_level_min_length_covering}:
\\
(i) returns $(l^*, u^*)$ that guarantees to exactly solve Problem (\ref{eq:instance_coverage_problem}), 
i.e., $(l^*, u^*) \in \arg\min_{(l, u) \in \calU(X; \tau)} \ell(l, u)$, 
and
\\
(ii) runs in $O(K)$ time complexity.
\end{theorem}

\noindent
We make three remarks on the above result for Algorithm \ref{alg:instance_level_min_length_covering}.

\begin{itemize}
\item 
\textbf{Model–agnostic.}  
The guarantee holds for \emph{any} predictive distribution $f(X)$, 
e.g., unimodal, multimodal, or otherwise.
No structural assumption on the network $f$ or the probability shape is required.

\item 
\textbf{Exact optimality.}  
Algorithm~\ref{alg:instance_level_min_length_covering} returns the \emph{exact} minimizer of Problem~\eqref{eq:instance_coverage_problem}.
In particular, the interval length $u^{*}-l^{*}$ is provably minimal among all intervals that satisfy the coverage and anchoring constraints.

\item
\textbf{Linear-time search.}  
Although the search space contains ${K\choose 2}+K=O(K^{2})$ candidate pairs $(l,u)$, the sliding–window procedure examines each boundary at most once, yielding an $O(K)$ run-time complexity.

\end{itemize}







  






\begin{algorithm}[t]
\caption{ min-CPS: Calibrating $\tau$ via Binary Search }
\label{alg:min_CPS}
\begin{algorithmic}[1]
\REQUIRE 
Nominal miscoverage rate $\alpha$,
a search boundary for $\tau$: $\underline \tau$ (lower) and $\overline \tau$ (upper)

\STATE Initialize $\tau \gets (\underline \tau + \overline \tau) / 2$

\FOR{$t=1, ..., T$}

  \STATE $\{(l^*(X_i; \tau), u^*(X_i; \tau))\}_{i=1}^n \gets$ Algorithm \ref{alg:instance_level_min_length_covering} on $D_\cal$

  \STATE Compute $F(\tau)$ as per (\ref{eq:empirical_coverage_rate})

  \STATE {\bf if} $F(\tau) \leq 1-\alpha$, {\bf then} $\overline \tau \gets \tau$

  \STATE {\bf else} $\underline \tau \gets \tau$

  \STATE $\tau \gets (\underline \tau + \overline \tau) / 2$

\ENDFOR

\RETURN Calibrated threshold $\tau$
\end{algorithmic}
\end{algorithm}

\subsection{ Minimum-Length Conformal Prediction Sets }
\label{subsection:min_CPS}

With the optimality guarantee of Algorithm \ref{alg:instance_level_min_length_covering} on each instance $(X, Y) \sim \calP$ given $\tau \in (0,1)$, 
we are ready to consider the search problem for $\tau$ in \eqref{eq:ordinal_cp_without_considering_predictive_efficiency}.
Let the empirical coverage rate on $D_\cal$ be denoted by
\begin{align}\label{eq:empirical_coverage_rate}
F(\tau) \triangleq \frac{1}{n} \sum_{i=1}^n \indicator[ l^*(X_i, \tau) \leq Y_i \leq u^*(X_i, \tau) ]
.
\end{align}


\begin{definition}
\label{definition:radial_monotonicity}
(Radial monotonicity)
A sequence of real numbers $\{a_k\}_{k=1}^K$ satisfies radial monotonicity if:

\noindent
(i) (Unique mode) there exists a unique index:
$$
m = \arg\max_{1\leq k\leq K} a_k ;
$$

\noindent
(ii) (Monotonicity in distance from the mode)
for any indices $k_1, k_2 \in [K]$,
$|k_1 - m| \leq |k_2 - m| \Rightarrow a_{k_1} \geq a_{k_2}$.
\end{definition}

\begin{lemma}
\label{lemma:nonconformity_score_satisfy_properties}
If $f(X)$ satisfies radial monotonicity for any $X$,
then the prediction sets constructed by min-CPS are nested in $\tau$: $\tau_1 \leq \tau_2 \Rightarrow \widehat C_{\tau_1}(X) \subseteq \widehat C_{\tau_2}(X)$.
Moreover, empirical coverage rate $F(\tau)$ is 
(i) monotonically non-decreasing in $\tau$, and
(ii) 
invariant to the orderings of calibration samples.
\end{lemma}

\begin{figure}[t]
    \centering
    \includegraphics[width=0.49\linewidth]{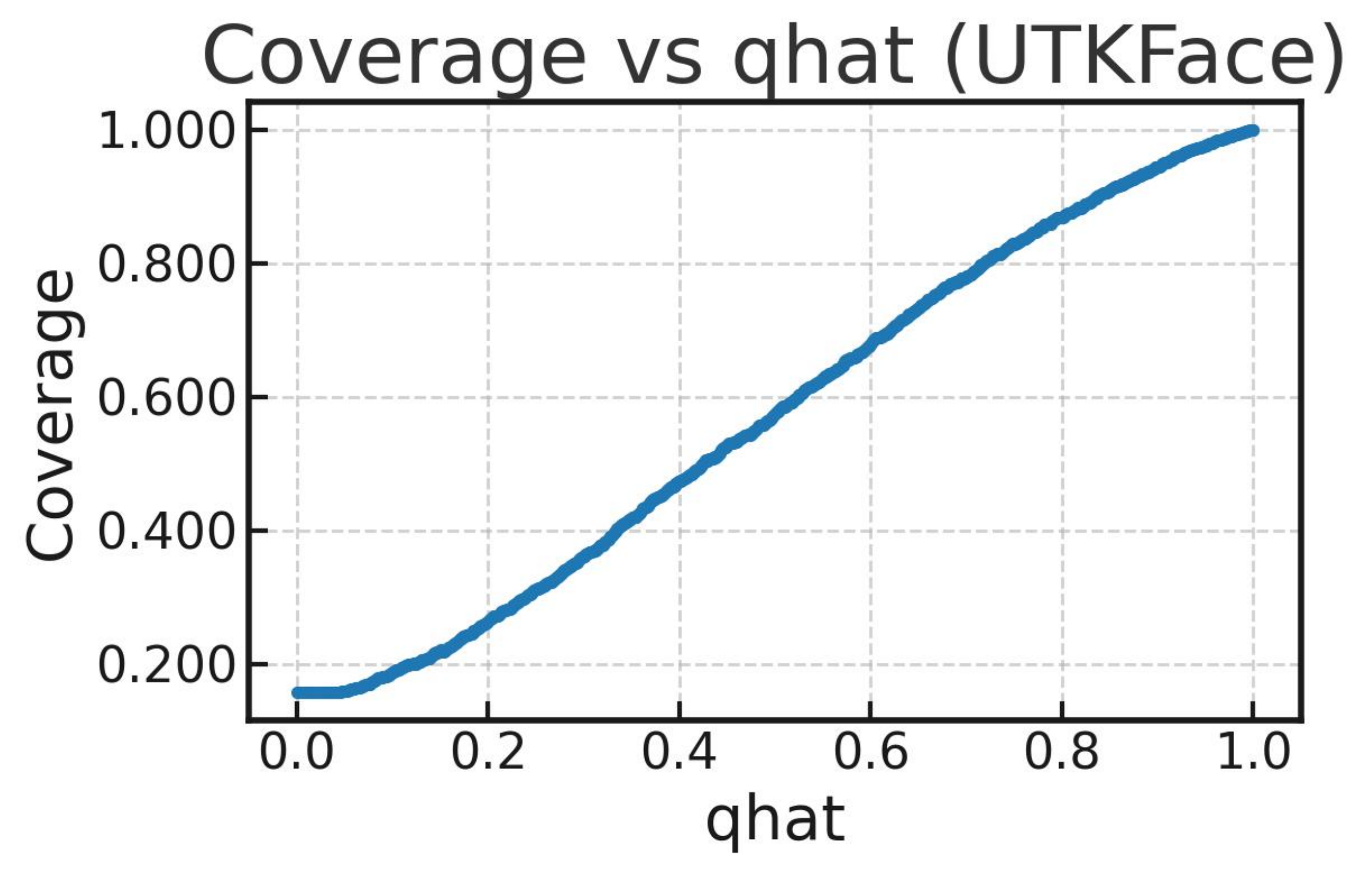}
    \includegraphics[width=0.49\linewidth]{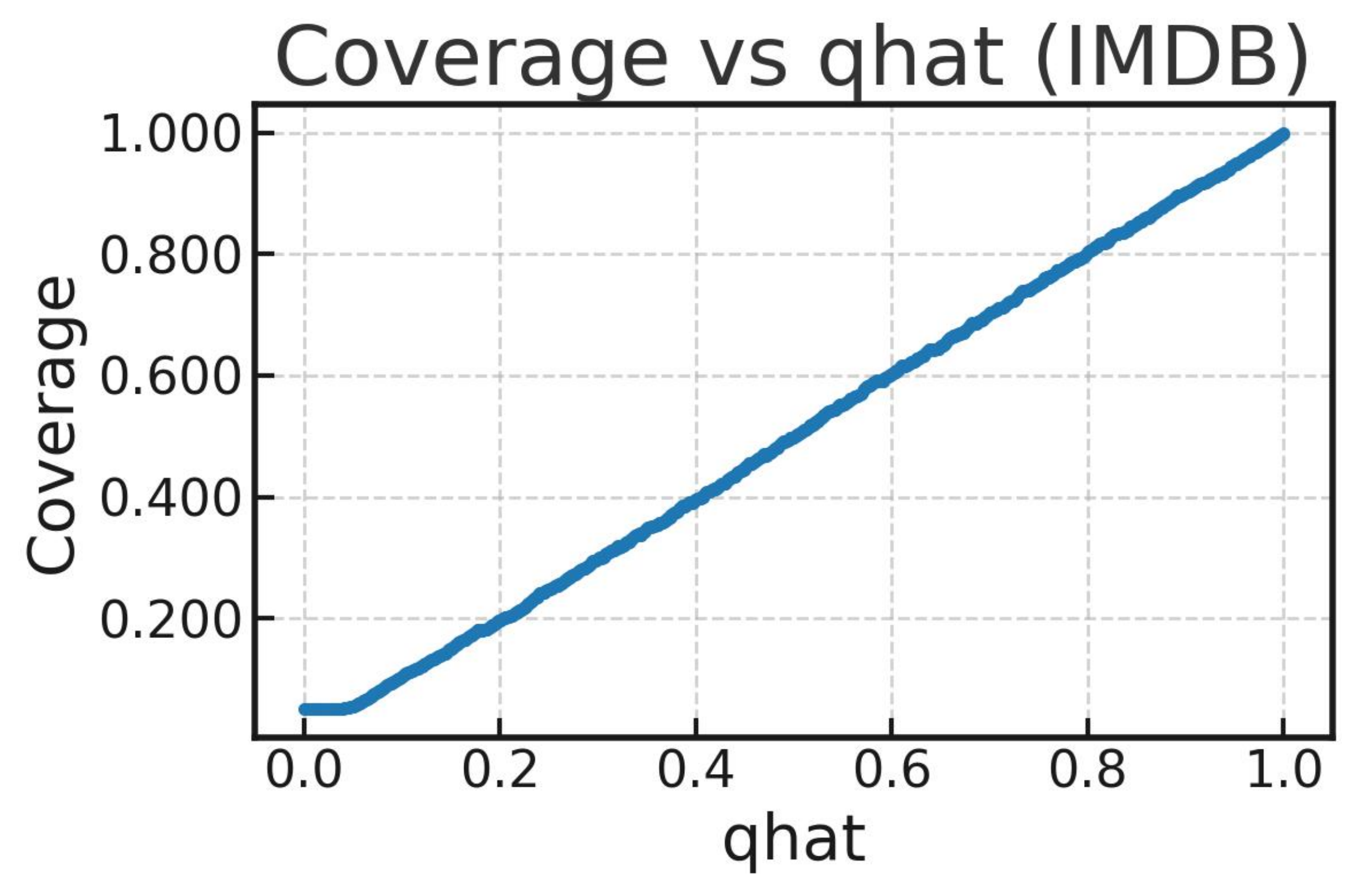}
    \caption{The empirical coverage rate $F(\tau)$: verify $F(\tau)$ monotonically increases in $\tau$ (ref. Section \ref{subsection:min_CPS}).
    }
    \label{fig:monotone_F}
\end{figure}

Given the monotonicity of $F(\tau)$ (verified in Figure \ref{fig:monotone_F}), binary search is a natural choice to find a sufficiently good value for $\tau$, such that $F(\tau) \geq 1-\alpha$.
We summarize this binary search procedure to determine $\tau$ in Algorithm \ref{alg:min_CPS},
which leads us to the following in-effect problem of min-CPS:
\begin{align}
\label{eq:ordinal_cp_with_considering_predictive_efficiency}
\min &~~ \tau ~~~~~~~ \st
\\
\sum_{i=1}^n & \indicator[ l^*(X_i; \tau) \leq Y_i \leq u^*(X_i; \tau) ] \geq \lceil ( 1 - \alpha ) ( n + 1 ) \rceil,
\nonumber\\
( l^*(&X_i; \tau), u^*(X_i; \tau) ) \in \arg\min_{ (l, u) \in \calU(X_i; \tau)} \ell(l, u), 
\forall i
.
\nonumber
\end{align}

Due to the monotonicity of $F(\tau)$, it converges to the optimal value of $\tau$ for coverage at the rate of $O(\exp(-T))$.
In other words, to achieve an $\epsilon$-optimal threshold $\tau$, 
the time complexity for min-CPS is $O(\log(1/\epsilon) n K)$, which is practically efficient (ref. running time comparison in Table \ref{tab:running_time}).

\begin{theorem}
\label{theorem:minCPS_coverage_guarantee}
(Coverage guarantee of min-CPS)
Under the same radial monotonicity assumption of Lemma \ref{lemma:nonconformity_score_satisfy_properties},
the calibrated threshold 
$\tau$ determined by Algorithm \ref{alg:min_CPS} yields the ($1-\alpha$) marginal coverage guarantee as in (\ref{eq:marginal_coverage_in_ordinal_classification}).
\end{theorem}



\paragraph{Comparison Problem (\ref{eq:ordinal_cp_with_considering_predictive_efficiency}) with Problem (\ref{eq:ordinal_cp_without_considering_predictive_efficiency}).}
As aforementioned, the standard formulation of ordinal CP in (\ref{eq:ordinal_cp_without_considering_predictive_efficiency}) does not necessarily take the predictive efficiency into account of the objective.
Consequently, there is no guarantee of close to smallest intervals and it is unclear how to approach the optimal predictive efficiency.
In contrast, by min-CPS, we solve an effective problem in (\ref{eq:ordinal_cp_with_considering_predictive_efficiency}),
which indeed minimizes the interval length on each instance,
and guarantees the superior predicitve efficiency of min-CPS in practice.

\subsection{ Length-Regularization Variant: Min-RCPS }
\label{subsection:min_RCPS}

In addition to min-CPS,
in this subsection,
we further consider the semantic coherence of prediction intervals and propose a length-aware regularization into min-CPS to improve the predictive efficiency.
Our motivation stems from the intuition that examples with larger minimum interval length $\ell(l^*(X; \tau), u^*(X; \tau))$ should be more uncertain than those with smaller minimum interval length,
so we need to penalize the data with larger length via reducing their cumulative probability by a certainty quantity depending on their length.

Formally, we redefine the cumulative probability used in the feasible constraint of the instance-level minimum-length covering problem in \eqref{eq:instance_coverage_problem} as follows:
\begin{equation}\label{eq:redefine_uncertainty_sets}
\calU_{\lambda}(X; \tau) 
\triangleq
\left\{
    (l, u) :
    \begin{aligned}
        &\sum_{k = l}^{u} f(X)_k \underbrace{ - \lambda \cdot \ell(l, u) }_{ \text{length reg.} } \geq \tau, \\
        &l \leq \hat{y}^*(X) \leq u
    \end{aligned}
\right\}
,
\end{equation}
where we simply penalize the cumulative probability by a linear term of $\ell(l, u)$ scaled by a hyper-parameter $\lambda$.
Beyond this redefinition of the feasible set $\calU_\lambda(X; \tau)$, 
we keep everything else of min-CPS as it is, 
including solving the instance-level problem \eqref{eq:instance_coverage_problem}, 
and the binary search for $\tau$ as in Algorithm \ref{alg:min_CPS}.
We refer this algorithmic procedure as minimum-length regularized conformal prediction sets (min-RCPS).
Analogous to min-CPS in \eqref{eq:ordinal_cp_with_considering_predictive_efficiency}, we also derive the objective of min-RCPS by using $\calU_\lambda$ of \eqref{eq:redefine_uncertainty_sets} as follows:
\begin{align}
\label{eq:ordinal_cp_with_considering_predictive_efficiency_regularized}
\min &~~ \tau ~~~~~~~ \st
\\
\sum_{i=1}^n & \indicator[ l^*(X_i; \tau) \leq Y_i \leq u^*(X_i; \tau) ] \geq \lceil ( 1 - \alpha ) ( n + 1 ) \rceil,
\nonumber\\
( l^*_\lambda(&X_i; \tau), u^*_\lambda(X_i; \tau) ) \in \arg\min_{ (l, u) \in \calU_\lambda(X_i; \tau)} \ell(l, u), 
\forall i
.
\nonumber
\end{align}
Notably, when $\lambda = 0$, the regularization term vanishes and min-RCPS reduces min-CPS.
We highlight that min-RCPS does not break the assumption of data exchangeability, so we have the following result:
\begin{corollary}
\label{corollary:regularization_improves_predictive_efficiency}
(Marginal coverage of min-RCPS)
Min-RCPS preserves exchangeability and follows the same split-conformal calibration rule as min-CPS.
Therefore, the calibrated threshold $\hat \tau$ ensures the standard marginal coverage guarantee as in (\ref{eq:marginal_coverage_in_ordinal_classification}).
\end{corollary}

\section{Experiments }
\label{section:experiments}


\subsection{Experimental Setup}
\label{subsection:Experimental_Setup}

\paragraph{Overview.}
We conduct extensive experiments to demonstrate the effectiveness of our proposed method on several real-world datasets. 
Following the literature of CP, 
we specifically evaluate two crucial metrics of ordinal CP methods: coverage and prediction set size.
We compare our results against the state-of-the-art baseline Ordinal APS \cite{lu2022ordinalaps}, a na\"ive method (referred to as Naive CDF, which is first used in \cite{lu2022ordinalaps} and a recent method WCRC \cite{xu2023crcoc} as a baseline).

\subsubsection{Datasets.}

We evaluate our method on four publicly-available benchmark datasets for ordinal classification, 
which demonstrate both the modality diversity and the real-world relevance of ordinal classification:

\begin{itemize}
  \item \textbf{UTKFace (Kaggle version)}~\cite{utkface_kaggle}: A large‐scale facial age dataset with over 20{,}000 images. Age is an inherently \emph{ordered} variable, making it an ideal test-bed for ordinal classification methods. 

  \item \textbf{Avocado Price}~\cite{avocado_kaggle}: Historical weekly avocado prices discretised into ordinal price bands. 
  This tabular, time-series dataset represents economic or financial forecasting scenarios where the outcome (price level) follows a natural ranking.
  It tests whether our approach generalises beyond vision data to structured numerical features subject to temporal trends and market volatility.  

  \item \textbf{Electric Motor Temperature}~\cite{motor_temp_kaggle}: Multivariate sensor readings for monitoring motor temperature, discretised into ordinal risk/temperature ranges. 
  Industrial condition-monitoring is a safety-critical domain: well-calibrated uncertainty is crucial to avoid both false alarms and missed overheating events. 
  The sequential nature and class imbalance pose additional challenges for ordinal coverage guarantees.  
  \item \textbf{IMDB}~\cite{Rothe-ICCVW-2015}: A large-scale facial age dataset containing over 500,000 images collected from IMDB and Wikipedia. 
  The dataset provides a challenging test-bed for ordinal classification and uncertainty quantification methods. Its scale, diverse real-world face conditions, and long-tailed age distribution make it a widely adopted benchmark in recent works \cite{keramati2024conrcontrastiveregularizerdeep, dong2025improverepresentationimbalancedregression}.

\end{itemize}

Together, these datasets span \emph{image}, \emph{tabular-economic}, and \emph{time-series sensor} modalities, enabling us to assess the robustness and general applicability of our conformal ordinal predictors across heterogeneous real-world tasks.

\begin{table}[t]
\centering
\caption{
{\bf Coverage and prediction set size ($\alpha=0.1$)}.
Bold numbers indicate the smallest prediction set size, while all methods guarantees at least $1-\alpha$ coverage rate. 
On Temperature, UTKFace, Avocado Price and IMDB, comparison with the best baseline (Ordinal APS), 
{\bf min-CPS reduces the prediction set size by} $5.11\%${$\downarrow$}, $7\%${$\downarrow$}, $39.93\%${$\downarrow$}, $3.49\%${$\downarrow$},
while {\bf min-RCPS reduces it by} $6.96\%${$\downarrow$}, $7.01\%${$\downarrow$}, $41.21\%${$\downarrow$}, $3.16\%${$\downarrow$}.
{\bf Average over all datasets, min-CPS and min-RCPS reduce the prediction set size by} $14\%${$\downarrow$} and $15\%${$\downarrow$}.
} 
\resizebox{\columnwidth}{!}{
\begin{tabular}{llcc}
\toprule
\textbf{Dataset} & \textbf{Method} & \textbf{Coverage} & \textbf{Prediction Set Size} \\
\midrule

\multirow{5}{*}{Temperature}
& Naive CDF     & 0.9002 $\pm$ 0.0041  & 41.2510 $\pm$ 0.3770 \\
& Ordinal APS   & 0.9015 $\pm$ 0.0040  & 5.3645 $\pm $ 0.0609 \\
& WCRC          & 0.9025 $\pm$ 0.0030  & 5.3538 $\pm $ 5.0566 \\
& min-CPS       & 0.9017 $\pm$ 0.0048  & 5.0902 $\pm $ 0.0581 \\ 
& min-RCPS      & 0.9021 $\pm$ 0.0034  & \textbf{ 4.9913 $\pm $ 0.0428} \\
\midrule

\multirow{5}{*}{UTKFace}
& Naive CDF     & 0.9025 $\pm$ 0.0048 & 73.6699 $\pm$ 0.9021 \\
& Ordinal APS   & 0.9021 $\pm$ 0.0071 & 32.0085 $\pm$ 0.3504 \\
& WCRC          & 0.9008 $\pm$ 0.0046 & 31.7998 $\pm $ 18.3118  \\
& min-CPS       & 0.9026 $\pm$ 0.0055 & 29.7669  $\pm$ 0.2332 \\ 
& min-RCPS      & 0.9025 $\pm$ 0.0055 & \textbf{ 29.7661 $\pm$ 0.2094} \\
\midrule

\multirow{5}{*}{Avocado Price}
& Naive CDF     & 0.9012 $\pm$ 0.0046 & 23.2533 $\pm$ 0.0693 \\
& Ordinal APS   & 0.9024 $\pm$ 0.0030 & 15.1795 $\pm$ 0.0596 \\
& WCRC          & 0.9014 $\pm$ 0.0043 & 15.1587 $\pm $ 4.9565 \\
& min-CPS       & 0.9034 $\pm$ 0.0057 & 9.1179  $\pm$ 0.0964 \\
& min-RCPS      & 0.9033 $\pm$ 0.0054 & \textbf{ 8.9235 $\pm$ 0.0510 } \\
\midrule

\multirow{5}{*}{IMDB}
& Naive CDF     & 0.8984 $\pm$ 0.0062 & 37.3994 $\pm$ 0.6318 \\
& Ordinal APS   & 0.9022 $\pm$ 0.0064 & 29.0753 $\pm$ 0.3479 \\
& WCRC          & 0.9008 $\pm$ 0.0000 & 29.1937 $\pm $ 0.0122\\
& min-CPS       & 0.9022 $\pm$ 0.0055 & \textbf{ 28.1761  $\pm$ 0.3339 }\\ 
& min-RCPS      & 0.9019 $\pm$ 0.0072 & 28.2635 $\pm$ 0.3791 \\

\bottomrule
\end{tabular}
}
\label{tab:coverage_set_size}
\end{table}

\subsubsection{Baselines.}
We consider three leading ordinal conformal prediction methods as baselines:

\begin{itemize}[leftmargin=*,topsep=1pt,itemsep=1pt]
\item \textbf{Ordinal APS} \cite{lu2022ordinalaps}: A variant of APS specifically tailored for ordinal classification, ensuring contiguous intervals.
\item \textbf{Naive CDF} (also used in \cite{lu2022ordinalaps} as a baseline): 
A simple method to construct prediction intervals directly from cumulative probabilities.
\item \textbf{Weighted CRC} \cite{xu2023crcoc}: A conformal risk-control method for ordinal classification that weights miscoverage by class-dependent severity to produce calibrated interval predictions.
\end{itemize}

\subsubsection{Evaluation Metrics.}
The evaluation metrics used in our experiments include the coverage of the true label and the average size of prediction sets under coverage,
which are detailed as follows:
\begin{itemize}
\item 
\textbf{Coverage (examine if $\geq 1-\alpha$)} is defined as the empirical coverage rate and can be computed by:
\begin{align*}
\text{Coverage} = \frac{1}{n}\sum_{i=n+1}^{n+m} \indicator [ Y_i \in \widehat C(X_i) ]
,
\end{align*}
where $D_\test = \{ (X_i, Y_i) \}_{i=n+1}^{n+m}$ is the test dataset with $m$ being the number of test samples.
$\widehat C$ denotes any ordinal CP methods (i.e., Naive CDF, Ordinal APS, min-CPS or min-RCPS) used in the experiment.

\item 
\textbf{The average size of prediction sets (smaller better)} is defined as average number of classes in prediction sets:
\begin{align*}
\text{Prediction set size}
=
\frac{1}{m} \sum_{i=n+1}^{n+m} \sum_{y\in\calY} \indicator[ y \in \widehat C(X_i) ] 
,
\end{align*}
which iterates all test samples and counts how many ordinal labels are included in the $\widehat C(X_i)$.
Smaller sets imply higher prediction certainty and informativeness.
\end{itemize}

\paragraph{Parameter Setting.}
We perform experiments under different miscoverage rates $\alpha$ (i.e., 0.1, 0.05, and 0.01) to thoroughly evaluate performance and robustness.
For min-RCPS, we tune the length regularization hyper-parameter $\lambda$ from a reasonable range $\{ 0, 0.001, 0.003, 0.005, ..., 0.019\}$ (with $0.002$ increasing step from $0.001$),
where we also conduct a sensitivity experiment later.

\subsection{ Experiment Results }
\label{subsection:experiment_result}

\begin{table}[t]
\centering
\caption{
{\bf Coverage and prediction set size on Avocado Price with various} $\boldsymbol{\alpha}$. 
(1) When $\alpha = 0.1$, the prediction set size decreases by $39.93\%${$\downarrow$} for min-CPS and 41.21\%{$\downarrow$} for min-RCPS.
(2) When $\alpha = 0.05$, min-CPS and min-RCPS reduce the prediction set size by 36.22\%{$\downarrow$} and 35.58\%{$\downarrow$}, respectively. 
(3) When $\alpha = 0.01$, the reduction is 24.80\%{$\downarrow$} for min-CPS and 26.38\%{$\downarrow$} for min-RCPS. 
On average across all three $\alpha$ values, min-CPS and min-RCPS reduce the prediction set size by 33.65\%{$\downarrow$} and 34.39\%{$\downarrow$}, respectively.
} 
\resizebox{\columnwidth}{!}{
\begin{tabular}{llcc}
\toprule
$\boldsymbol{\alpha}$ & \textbf{Method} & \textbf{Coverage} & \textbf{Prediction Set Size} \\
\midrule

\multirow{5}{*}{$\alpha = 0.1$}
& Naive CDF     & 0.9012 $\pm$ 0.0046 &  23.2533 $\pm$ 0.0693 \\
& Ordinal APS   & 0.9024 $\pm$ 0.0030 &  15.1795 $\pm$ 0.0596 \\
& WCRC          & 0.9014 $\pm$ 0.0043 & 15.1587 $\pm $ 4.9565\\
& min-CPS       & 0.9034 $\pm$ 0.0057 & 9.1179  $\pm$ 0.0964 \\ 
& min-RCPS      & 0.9033 $\pm$ 0.0054 & \textbf{ 8.9235 $\pm$ 0.0510 } \\
\midrule

\multirow{4}{*}{$\alpha = 0.05$}
& Naive CDF     & 0.9507 $\pm$ 0.0042 & 25.5960 $\pm$ 0.1128 \\
& Ordinal APS   & 0.9509 $\pm$ 0.0026 & 16.7082 $\pm$ 0.0648 \\
& WCRC          & 0.9520 $\pm$ 0.0000 & 16.6785 $\pm $ 0.0523\\
& min-CPS       & 0.9517 $\pm$ 0.0036 & \textbf{10.6557 $\pm$ 0.0959} \\ 
& min-RCPS      & 0.9516 $\pm$ 0.0036 & 10.7631 $\pm$ 0.0733  \\ 
\midrule

\multirow{4}{*}{$\alpha = 0.01$}
& Naive CDF     & 0.9904 $\pm$ 0.0019 & 29.2091 $\pm$ 0.1967 \\
& Ordinal APS   & 0.9903 $\pm$ 0.0018 & 19.6791 $\pm$ 0.1949 \\
& WCRC          & 0.9905 $\pm$ 0.0000 & 19.7776 $\pm $ 0.0216\\
& min-CPS       & 0.9901 $\pm$ 0.0016 & 14.7986 $\pm$ 0.2855  \\ 
& min-RCPS      & 0.9894 $\pm$ 0.0019 & \textbf{14.4887 $\pm$ 0.3399} \\ 

\bottomrule
\end{tabular}
}
\label{tab:coverage_different_alpha}
\end{table}

\subsubsection{Comparison with Baselines Across Benchmark Datasets.}
The overall performance comparison of all ordinal CP methods across three benchmark datasets is reported in Table~\ref{tab:coverage_set_size},
where we follow a standard practice in the CP literature \cite{romano2020classification,angelopoulos2020uncertainty,huang2024labelrank} and set $\alpha=0.1$.
The results clearly demonstrate that both min-CPS and min-RCPS significantly improve the predictive efficiency over the best one of Ordinal APS, Naive CDF and WCRC.

Specifically, on Temperature, UTKFace, Avocado Price and IMDB,
min-CPS relatively reduces the prediction set size by $5.11\%${$\downarrow$}, $7\%${$\downarrow$}, $39.93\%${$\downarrow$} and $3.49\%${$\downarrow$},
while min-RCPS relatively reduces the prediction set size by $6.96\%${$\downarrow$}, $7.01\%${$\downarrow$}, $41.21\%${$\downarrow$} and $3.16\%${$\downarrow$}.
On average across all three datasets, min-CPS and min-RCPS reduce the prediction set size by $14\%${$\downarrow$} and $15\%${$\downarrow$}, respectively.
These relative improvements of predictive efficiency (more than $10\%\downarrow$ in prediction set size) are clearly significant over the CP literature \cite{ding2024class,huang2024labelrank,shi2024conformal}.

\subsubsection{Sensitivity Analysis under Different Miscoverage Levels.}
We further evaluate the methods under varying miscoverage levels, i.e., $\alpha \in \{ 0.1, 0.05, 0.01 \}$,
on the UTKFace dataset.
We report the results of coverage and prediction set size for all methods in Table~\ref{tab:coverage_different_alpha}.
Our results indicate that our methods consistently achieves smaller prediction sets across all evaluated $\alpha$ values, highlighting its stability and robustness.

Specifically, our observation can be summarized by
\begin{itemize}
\item 
When $\alpha = 0.1$, the prediction set size decreases by $39.93\%${$\downarrow$} for min-CPS and 41.21\%{$\downarrow$} for min-RCPS.

\item 
When $\alpha = 0.05$, min-CPS and min-RCPS reduce the prediction set size by 36.22\%{$\downarrow$} and 35.58\%{$\downarrow$}, respectively, compared with the best baseline. 

\item 
When $\alpha = 0.01$, the reduction is 24.80\%{$\downarrow$} for min-CPS and 26.38\%{$\downarrow$} for min-RCPS. 
On average across all three $\alpha$ values, min-CPS and min-RCPS reduce the prediction set size by 33.65\%{$\downarrow$} and 34.39\%{$\downarrow$}, respectively.
\end{itemize}

\begin{figure}[t]
    \centering
    \includegraphics[width=0.49\linewidth]{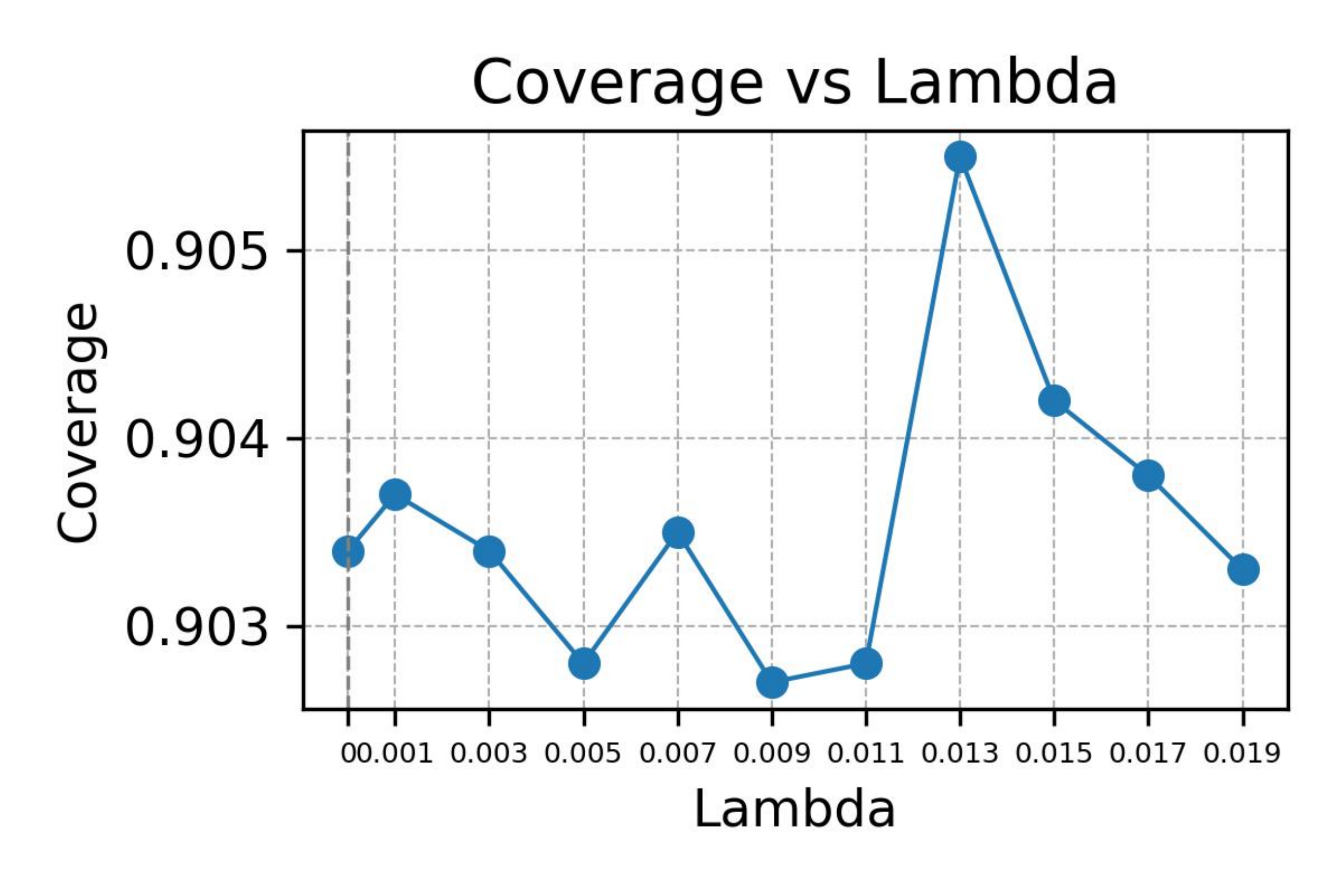}
    \includegraphics[width=0.49\linewidth]{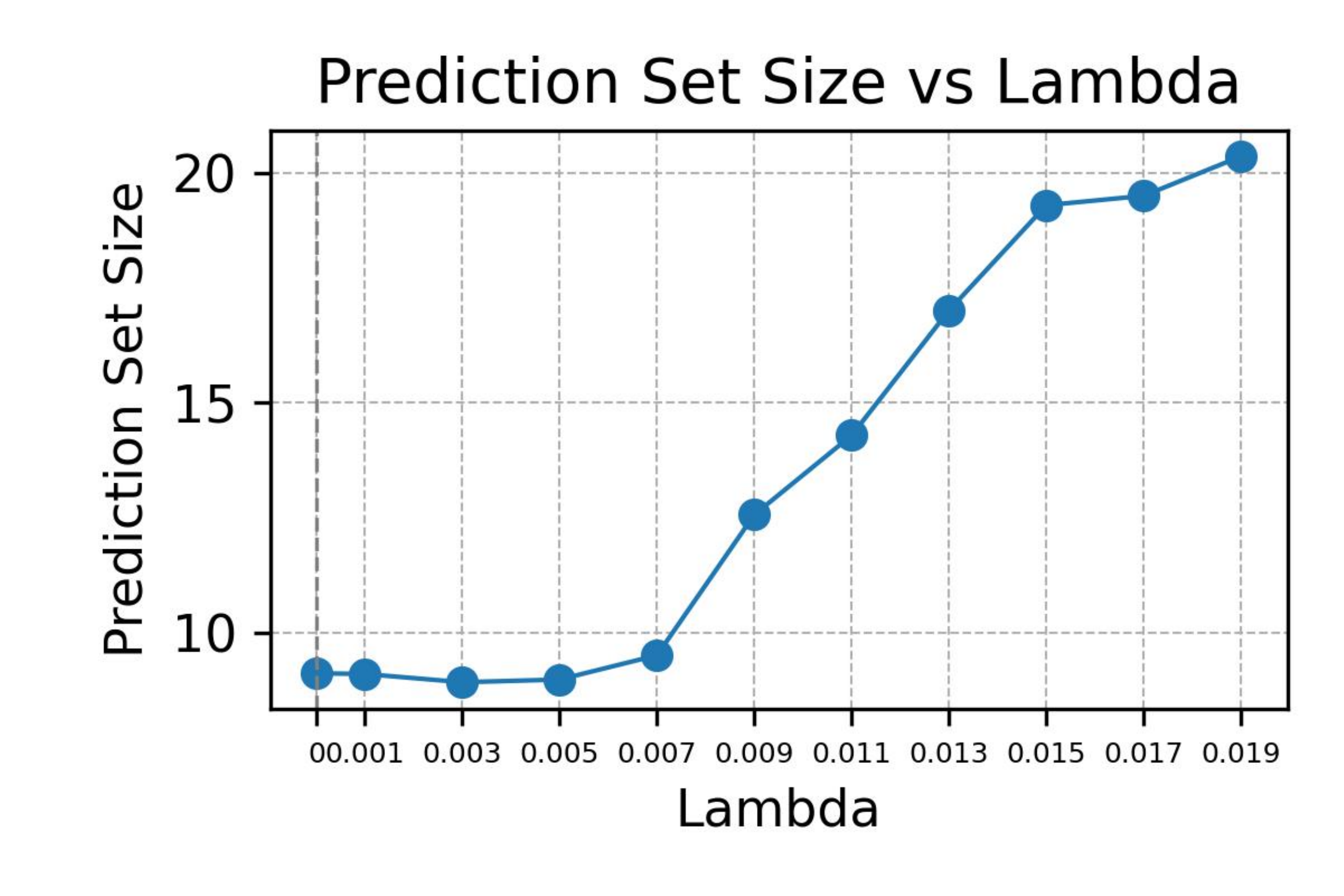}
    \caption{Impact of the regularization hyper-parameter $\lambda$ in min-RCPS on coverage and prediction set size on the Avocado Price dataset.
    When the coverage is guaranteed, the prediction set size initially decreases as $\lambda$ increases from $0$ to 0.003,
    where min-RCPS outperforms min-CPS by 2.13\%{$\downarrow$} in terms of prediction set size.
    We find that setting $\lambda$ to a relatively small value is sufficient to improve the predictive efficiency, and this happens on other datasets.
    }
    \label{fig:lambda_curves}
\end{figure}

\subsubsection{Impact of Regularization Parameter $\lambda$ in min-RCPS.}
We analyze the effect of the regularization hyperparameter $\lambda$ for min-RCPS on coverage and prediction set sizes. 
Figures~\ref{fig:lambda_curves} illustrate these effects on the Avocado Price datasets:
\begin{itemize}[leftmargin=*,topsep=1pt,itemsep=1pt]
\item As $\lambda$ increases from $0$, the prediction set sizes initially decrease and reach a smallest prediction set size when $\lambda = 0.003$.
\item Appropriately choosing $\lambda$ significantly enhances the compactness of prediction sets as maintaining valid coverage.
\item We recommend selecting the optimal $\lambda$ via cross-validation or a calibration set for practical applications.
\end{itemize}

\subsubsection{Running time comparison.}
We also compare the running time between our methods and baselines. Table~\ref {tab:running_time} presents the running time comparison across three datasets under the setting $\alpha = 0.1$ with 10 trials. Our proposed methods, min-CPS and min-RCPS, demonstrate a significant computational advantage over the baseline Ordinal APS. Specifically, on Temperature, UTKFace, Avocado Price and IMDB,
min-CPS speeds up x19.82, x21.77, x10.62 and x9.80 over Ordinal APS,
while min-RCPS speeds up x19.26, x22.58, x11.29 and x9.46 over Ordinal APS. For the fair comparison, the stop criteria for both min-CPS and min-RCPS are to keep iterating until the coverage has been guaranteed (exceeding $1-\alpha$).

\begin{table}[t]
\centering
\caption{{\bf Running time across datasets ($\alpha=0.1$)}.
All methods guarantee at least $1 - \alpha$ coverage rate with 10 trials. On Temperature, UTKFace, Avocado Price and IMDB,
{\bf min-CPS} speeds up x19.82, x21.77, x10.62 and x9.80 over Ordinal APS,
while {\bf min-RCPS} speeds up x19.26, x22.58, x11.29 and x9.46 over Ordinal APS. The stop criteria for both min-CPS and min-RCPS are to keep iterating until the coverage has been guaranteed (exceeding $1-\alpha$).
}
\resizebox{.9\columnwidth}{!}{
\begin{tabular}{llcc}
\toprule
\textbf{Dataset} & \textbf{Method} & \textbf{Running Time (s)} \\
\midrule

\multirow{4}{*}{Temperature}
& Ordinal APS   &  6183  \\
& min-CPS       &  312 \\ 
& min-RCPS      &  321  \\
\midrule

\multirow{4}{*}{UTKFace}
& Ordinal APS   & 4855 \\
& min-CPS       & 223 \\ 
& min-RCPS      & 215 \\
\midrule

\multirow{4}{*}{Avocado Price}
& Ordinal APS   & 711 \\
& min-CPS       & 67 \\
& min-RCPS      & 63 \\
\midrule

\multirow{4}{*}{IMDB}
& Ordinal APS   & 823 \\
& min-CPS       & 84 \\
& min-RCPS      & 87 \\
\bottomrule
\end{tabular}
}
\label{tab:running_time}
\end{table}

\section{Conclusion}
\label{section:conclusion}

In this work, we introduced min-CPS, a model-agnostic conformal prediction method for ordinal classification based on an instance-level minimum-length covering formulation. A linear-time sliding-window algorithm is developed to solve this problem exactly. Under a mild radial monotonicity condition, we show that the resulting prediction sets are nested in the calibration threshold, which ensures the monotonicity of the empirical coverage and enables efficient calibration with valid marginal coverage.
We further proposed min-RCPS, a length-regularized extension that incorporates interval-length information while preserving the same coverage guarantee through standard split-conformal arguments.
Across vision, sensor, and tabular datasets, both methods maintain the target coverage and yield more compact prediction sets than existing ordinal CP approaches. min-CPS and min-RCPS reduce average interval sizes by 14\% and 15\%, with reductions up to 30–40\% in some cases, and achieve 10x–22x speedups over Ordinal APS.


\newpage

\section*{Acknowledgements}

The authors gratefully acknowledge the in part support by the USDA-NIFA funded AgAID Institute
award 2021-67021-35344, and the NSF grant CNS-2312125, IIS-2443828, DUE-2519063. 
The views expressed are those of
the authors and do not reflect the official policy or position of the USDA-NIFA and NSF.

\bibliography{aaai2026}

\input{appendix_aaai26_camera_ready_final}

\end{document}

%% file: appendix_aaai26_camera_ready_final.tex
\appendix

\onecolumn

\section{ Proofs for Main Results }
\label{section:proofs_for_main_results}

\subsection{ Proof for Theorem \ref{theorem:complexity_minimum_length_instance_coverage} }
\label{subsection:appendix:proof:theorem:complexity_minimum_length_instance_coverage}
\begin{theorem_appendix}
\label{theorem:appendix:complexity_minimum_length_instance_coverage}
(Theorem \ref{theorem:complexity_minimum_length_instance_coverage} restated, optimality and complexity of Algorithm \ref{alg:instance_level_min_length_covering})
Let $K \in \mathbb N$ and $\tau \in (0, 1]$.
For any input $X \in \calX$,
Algorithm \ref{alg:instance_level_min_length_covering}:
\\
(i) returns $(l^*, u^*)$ that guarantees to exactly solve Problem (\ref{eq:instance_coverage_problem}), 
i.e., $(l^*, u^*) \in \arg\min_{(l, u) \in \calU(X; \tau)} \ell(l, u)$, 
and
\\
(ii) runs in $O(K)$ time complexity.
\end{theorem_appendix}

\begin{proof}
(of Theorem \ref{theorem:complexity_minimum_length_instance_coverage} )

\noindent
\paragraph{(i) Optimality.} 

To prove the optimality of Algorithm \ref{alg:instance_level_min_length_covering}, we address two concerns, i.e., {\it constraint feasibility} and {\it the minimum length}, respectively.

\noindent
{\it Constraint feasibility:} anchor inclusion and sufficient probability coverage withint the interval.

For the anchor inclusion constraint, we restrict the outer iterator $u$ to the range from $\hat y^*$ to $K$,
while the inner iterator $l$ to the range from $1$ to $\hat y^*$.

For the sufficient probability coverage ($\geq \tau$) within the interval, we use the prefix sum array to construct a constant-time search table that is able to return range sum given $l$ and $u$.
For each fixed the outer iterator $u$, we only examine the lower bound $l$ that sufficiently covers more than $\tau$ probability (Line \ref{alg:instance_level_min_length_covering:line:inner_loop}).

\noindent
{\it Minimum length.}
When the iterator only stops at the values on which the above two constraints are feasible, 
we record the achieved length, and then derive the smallest one to reach the optimality.

\noindent
\paragraph{(ii) Complexity.}

Our Algorithm \ref{alg:instance_level_min_length_covering} consists of two main parts in terms of computation.
First is to compute the {\it prefix sum array} (Line \ref{alg:instance_level_min_length_covering:line:prefix_sum_start} - \ref{alg:instance_level_min_length_covering:line:prefix_sum_end}) and the {\it main search loop} (starting from Line \ref{alg:instance_level_min_length_covering:line:outer_loop}).

For the {\it prefix sum array}, it takes $K$ iterations for $K$ class labels, which is determined by the application domain.
Apparently it takes $O(T)$ time complexity.

For the {\it main search loop}, even though there are two nested loops,
it only takes $O(K)$ iterations.
The reason is that
both the outer and inner iterators ($u$ and $l$, respectively) are only allowed to increase withing their domains, 
i.e., $u$ from $\hat y^*$ to $K$ and $l$ from 1 to $\hat y^*$.
More specifically, without these two iterators going back, we avoid all possible redundant searches that do not give a shorter length.
Therefore, combined all possible values of both iterators, there are only $K$ searches that we need to examine and finally figure out which one gives the exactly minimum length.
\end{proof}

\subsection{ Proof for Lemma \ref{lemma:nonconformity_score_satisfy_properties} }
\label{subsection:appendix:lemma:nonconformity_score_satisfy_properties}

\begin{lemma_appendix}
\label{lemma:appendix:nonconformity_score_satisfy_properties}
(Lemma \ref{lemma:nonconformity_score_satisfy_properties} restated)
If $f(X)$ satisfies radial monotonicity for any $X$,
then the prediction sets constructed by min-CPS are nested in $\tau$: $\tau_1 \leq \tau_2 \Rightarrow \widehat C_{\tau_1}(X) \subseteq \widehat C_{\tau_2}(X)$.
Moreover, empirical coverage rate $F(\tau)$ is 
(i) monotonically non-decreasing in $\tau$, and
(ii) 
invariant to the orderings of calibration samples.
\end{lemma_appendix}

\begin{proof}
(of Lemma \ref{lemma:nonconformity_score_satisfy_properties})

\paragraph{(1) Nested prediction sets.}

In order to prove the nestedness of the prediction sets constructed by min-CPS, we require the following technical lemma.
The proof for Lemma \ref{lemma:maximal_mass_among_length_L_intervals} is deferred to Section \ref{subsection:appendix:lemma:proof:maximal_mass_among_length_L_intervals}.

\begin{lemma}
\label{lemma:maximal_mass_among_length_L_intervals}
(Maximal mass among length-L intervals)
Let $\{ a_k \}_{k=1}^K$ satisfy radial monotonicity with unique mode $m$.
For each $L \geq 0$, let
\begin{align*}
S_L \triangleq \{m\} \cup \{ \text{the $L$ indices such that $k \neq m$ with smallest distance } d(k) \triangleq |k-m|, \text{ for } k \in [K] \} 
.
\end{align*}
Let $I_L \triangleq [\min S_L, \max S_L ]$ be the contiguous interval induced by $S_L$.
Then $I_L$ satisfies the following three properties:

(i) The length of $I_L$ is $L$, or equivalently, the cardinality of $I_L$ is $L+1$, i.e., $| I_L | = L+1$;

(ii) The interval $I_L$ contains the mode $m$, i.e., $m \in I_L$;

(iii) The mass summation covered by $I_L$ is greater than or equivalent to that of any intervals with length $L+1$, i.e., 
\begin{align*}
\sum_{k \in I_L} a_k 
=
\max_{ l \leq m \leq u, u-l=L} \sum_{k=l}^u a_k
.
\end{align*}
\end{lemma}





Fix any $X$.  For $\tau\in(0,1)$, denote by 
\[
L^*(X; \tau)=\min\{\,L\ge 0 : \max_{u-l=L,\; l\le \hat y^*(X)\le u}\sum_{k=l}^u f(X)_k \ge \tau \,\}
\]
the minimal interval length required to cover at least $\tau$ mass while including the anchor.
By Algorithm \ref{alg:instance_level_min_length_covering} and Lemma \ref{lemma:maximal_mass_among_length_L_intervals} (iii), the solution returned by Algorithm \ref{alg:instance_level_min_length_covering} is exactly
\begin{align}\label{eq:algorithm1_exactly_equivalent_to_I_L}
(l^*(X;\tau), 
u^*(X;\tau)) = I_{L^*(X; \tau)}
,
\end{align}
where $I_L$ is the maximal mass interval of length $L$ characterized in Lemma \ref{lemma:maximal_mass_among_length_L_intervals}.

\paragraph{Step 1.} $L^*(X; \tau)$ is non-decreasing in $\tau$.

If $\tau_1 \leq \tau_2$, any interval that covers mass $\geq \tau_2$ must also cover $\geq \tau_1$.
Therefore the minimal feasible length cannot decrease:
\begin{align}\label{eq:nested_L_in_tau}
\tau_1\le\tau_2 \;\Rightarrow\; L^*(X; \tau_1)\le L^*(X; \tau_2)
.
\end{align}

\paragraph{Step 2.}

The family $\{I_L\}_{L\ge 0}$ is nested in $L$.
By the construction in Lemma \ref{lemma:maximal_mass_among_length_L_intervals}, 
$S_L$ consists of the mode $m$ and the $L$ indices with the smallest distance to $m$.  
Thus, $S_L\subseteq S_{L+1}$ and consequently
\begin{align}\label{eq:nested_I_L}
I_L=[\min S_L,\max S_L]\subseteq[\min S_{L+1},\max S_{L+1}]=I_{L+1}.
\end{align}

\paragraph{Step 3.} 

Nestedness of intervals in $\tau$.
Combining (\ref{eq:algorithm1_exactly_equivalent_to_I_L}) – (\ref{eq:nested_I_L}), we have
\[
\tau_1\le\tau_2
\;\Rightarrow\;
L^*(X; \tau_1)\le L^*(X; \tau_2)
\;\Rightarrow\;
I_{L^*(X; \tau_1)}\subseteq I_{L^*(X; \tau_2)}
\;\Rightarrow\;
\widehat C_{\tau_1}(X)\subseteq \widehat C_{\tau_2}(X),
\]
which establishes the nestedness of the prediction sets.

\paragraph{(2) For the empirical coverage rate $F(\tau)$.}

\paragraph{(2.i) Monotonicity.}

Recall the definition of the empirical coverage rate $F(\tau)$ as follows (see \eqref{eq:empirical_coverage_rate}):
\begin{align*}
F(\tau) \triangleq \frac{1}{n} \sum_{i=1}^n \indicator[ l^*(X_i, \tau) \leq Y_i \leq u^*(X_i, \tau) ]
.
\end{align*}

Moreover, note that $(l^*(X), u^*(X))$ is determined by solving Problem \eqref{eq:instance_coverage_problem},
which we restated below:
\begin{align*}
&
\min_{(l, u) \in \calU(X; \tau)} ~ 
\underbrace{ \ell(l, u) \triangleq u - l
}_{ \text{interval length} },
~
\text{ \st }~
\tau \in (0, 1), \text{ and}
\\
&
\calU(X; \tau) 
= 
\Bigg\{
(l, u)
: 
\underbrace{
\sum_{k=l}^u f(X)_k 
}_{ \text{(i) covering prob.} }
\geq \tau,
\underbrace{
l \leq \hat y^*(X) \leq u 
}_{ \text{(ii) including anchor} }
\Bigg\}
.
\end{align*}

Due to the radial monotonicity, 
it is apparent to find that the length $\ell(l, u) = u - l$ is monotonically non-decreasing as $\tau$ gets larger,
due to the covering probability term $\sum_{k=l}^u f(X)_k \geq \tau$.
This monotonicity applies to any data $X$ uniformly,
so after marginalizing over $X$ from the entire distribution, such monotonicity still holds,
i.e., $F(\tau)$ also satisfies this monotonicity condition.

\paragraph{(2.ii) Permutation invariance to the orderings of calibration samples.}
Recall that this permutation invariance (or equivalently, data exchangeability) is one of requirements by CP \cite{vovk2005algorithmic,shafer2008tutorial} on which the coverage guarantee can possibly hold.

It is also apparent that $F(\tau)$ is invariant to the orderings of data, 
since it is simply a summation of coverage indicators over all $n$ samples.
If we exchange any part within this summation equation, the summation itself does not change,
so we $F(\tau)$ is invariant to the orderings of calibration samples.
\end{proof}

\subsection{ Proof for Theorem \ref{theorem:minCPS_coverage_guarantee} }
\label{subsection:appendix:proof:theorem:minCPS_coverage_guarantee}

\begin{theorem_appendix}
\label{theorem:appendix:minCPS_coverage_guarantee}
(Theorem \ref{theorem:minCPS_coverage_guarantee} restated, coverage guarantee of min-CPS)
Under the same radial monotonicity assumption of Lemma \ref{lemma:nonconformity_score_satisfy_properties},
the calibrated threshold 
$\tau$ determined by Algorithm \ref{alg:min_CPS} yields the ($1-\alpha$) marginal coverage guarantee as in (\ref{eq:marginal_coverage_in_ordinal_classification}).
\end{theorem_appendix}

\begin{proof}
(of Theorem \ref{theorem:minCPS_coverage_guarantee})

Fix any calibration set $D_{\text{cal}} = \{(X_i,Y_i)\}_{i=1}^n$ and a new test point $(X_{n+1},Y_{n+1})$.
For each $\tau\in(0,1)$, min-CPS constructs a prediction set
\[
\widehat C_\tau(X) = [\,l^*(X;\tau),\,u^*(X;\tau)\,]
\]
by solving Problem (3) with Algorithm 1. 

By Lemma \ref{lemma:nonconformity_score_satisfy_properties}, the family $\{\widehat C_\tau(X)\}_{\tau\in(0,1)}$ is:

\noindent
(i) nested in~$\tau$ in the sense that
\[
\tau_1 \le \tau_2 \;\Rightarrow\; \widehat C_{\tau_1}(X)\subseteq \widehat C_{\tau_2}(X),
\]
and 

\noindent
(ii) permutation–invariant with respect to the ordering of calibration samples.

Recall the empirical coverage rate function
\[
F(\tau)
= \frac{1}{n}\sum_{i=1}^n \indicator \{Y_i\in\widehat C_\tau(X_i)\}.
\]
By the nestedness property above, $F(\tau)$ is non–decreasing in $\tau$.
Algorithm~2 chooses a threshold $\widehat\tau$ such that
\begin{align}\label{eq:empirical_cover}
\sum_{i=1}^n \indicator\{Y_i\in\widehat C_{\widehat\tau}(X_i)\}
\;\ge\; \bigl\lceil (1-\alpha)(n+1)\bigr\rceil.
\end{align}

Now consider the $(n+1)$ exchangeable samples
$\{(X_i,Y_i)\}_{i=1}^{n+1}$.
Because the construction of $\widehat C_\tau$ is permutation–invariant
and depends on $D_{\text{cal}}$ only through the multiset of
indicator values $\mathbf{1}\{Y_i\in\widehat C_\tau(X_i)\}$,
the joint distribution of
\[
\bigl(\indicator\{Y_i\in\widehat C_{\widehat\tau}(X_i)\}\bigr)_{i=1}^{n+1}
\]
is exchangeable.
Consequently, under condition (\ref{eq:empirical_cover}), 
at most $\alpha(n+1)$ of these
$n+1$ indicators can be zero, so the probability that the test point
$(X_{n+1},Y_{n+1})$ is covered is at least
\[
\mathbb{P}\bigl\{Y_{n+1}\in\widehat C_{\widehat\tau}(X_{n+1})\bigr\}
\;\ge\; 1-\alpha.
\]

Rewriting this in the original notation yields
\[
\mathbb{P}_{X,Y}\bigl\{Y\in\widehat C_{\widehat\tau}(X)\bigr\}
\;\ge\; 1-\alpha,
\]
which is exactly the marginal coverage guarantee in (\ref{eq:marginal_coverage_in_ordinal_classification}).
\end{proof}

\subsection{ Proof for Corollary \ref{corollary:regularization_improves_predictive_efficiency} }
\label{subsection:appendix:proof_corollary:regularization_improves_predictive_efficiency}

\begin{corollary_appendix}
\label{corollary:appendix:regularization_improves_predictive_efficiency}
(Corollary \ref{corollary:regularization_improves_predictive_efficiency} restated, marginal coverage of min-RCPS)
Min-RCPS preserves exchangeability and follows the same split-conformal calibration rule as min-CPS.
Therefore, the calibrated threshold $\hat \tau$ ensures the standard marginal coverage guarantee as in (\ref{eq:marginal_coverage_in_ordinal_classification}).
\end{corollary_appendix}

\begin{proof}
(of Corollary \ref{corollary:regularization_improves_predictive_efficiency})

Consider the conformal predictor defined by min-RCPS,
which maps each $(X, \tau)$ to a set-valued prediction
$\widehat C_\tau^\lambda(X)$ as in Problem (\ref{eq:ordinal_cp_with_considering_predictive_efficiency_regularized}).
The construction is permutation–invariant with respect
to the ordering of the calibration samples, and the
$(n+1)$ pairs $\{(X_i,Y_i)\}_{i=1}^{n+1}$ are
exchangeable by assumption.

In particular, min-RCPS chooses a threshold $\hat\tau$
such that
\[
\sum_{i=1}^n \indicator\{Y_i \in \widehat C_{\hat\tau}^\lambda(X_i)\}
\;\ge\; \bigl\lceil(1-\alpha)(n+1)\bigr\rceil,
\]
which is the same calibration rule as in Theorem \ref{theorem:minCPS_coverage_guarantee}.
Therefore, the proof of Theorem \ref{theorem:minCPS_coverage_guarantee} applies verbatim
after replacing $\widehat C_\tau$ by $\widehat C_\tau^\lambda$,
and we obtain
\[
\mathbb{P}\{Y \in \widehat C_{\hat\tau}^\lambda(X)\}
\;\ge\; 1-\alpha.
\]
Moreover, the empirical coverage rate is clearly invariant
to any permutation of the calibration samples, since it is
a simple average of coverage indicators.
\hfill$\square$

The length-regularized variant min-RCPS inherits everything regarding CP guarantees from the original version min-CPS.
It only replaces how conformity score is computed, without changing any underlying assumptions, including permutation invariant, nestedness, etc.
As a result, it is apparent to see min-RCPS is also permutation-free and guarantee the same coverage, 
simply as in min-CPS.
\end{proof}

\subsection{ Proof for Lemma \ref{lemma:maximal_mass_among_length_L_intervals} }
\label{subsection:appendix:lemma:proof:maximal_mass_among_length_L_intervals}

\begin{proof}
(Lemma \ref{lemma:maximal_mass_among_length_L_intervals})

\paragraph{(i) Length $L$.}

Due to the radial monotonicity, the $L$ elements in $\{a_k\}_{k=1}^K$ with the smallest $d(k)$ (except the mode $m$) are actually the $L$ largest ones (except the mode $m$, 
so $I_L$ is continuous and includes $L+1$ elements, with $L$ length.

\paragraph{(ii) Containing $m$.}

By the construction of $S_L$, it trivially includes $m$.

\paragraph{(iii) Maximal mass summation.}

Suppose $T = \{l, ..., u\}$ is an interval with length $L$.
Due to the radial monotonicity, 
by pairing each index in $T \backslash S_L$ with an index in $S_L\backslash T$ that has no larger distance to $m$,
we have
\begin{align}\label{eq:sum_gap}
\sum_{k \in S_L \backslash T} a_k
\geq 
\sum_{k \in T \backslash S_L} a_k
.
\end{align}

Further, 
we have
\begin{align*}
\sum_{k \in S_L} a_k
- \sum_{k \in T} a_k
=
\sum_{k \in S_L \backslash T} a_k
-
\sum_{k \in T \backslash S_L} a_k
\geq 
0
,
\end{align*}
where the above inequality is due to (\ref{eq:sum_gap}),
which means that the mass summation covered by $I_L$ is maximal over all intervals with same length $L$.
\end{proof}

\section{Additional Experimental Details}

\subsection{Datasets and Ordinal Structure}

We evaluate our method on three real-world datasets with naturally ordered labels:

\textbf{Temperature Dataset.} This dataset contains continuous temperature readings over time, where the task is to predict the precise temperature value with a resolution of \textbf{0.01 degrees Celsius}. To adapt this into an ordinal classification task, we discretize temperature into evenly spaced bins, such that each class corresponds to a distinct 0.01°C interval. This transforms the task into predicting from a set of finely grained and inherently ordered temperature levels, where class labels maintain the natural monotonic relationship of physical temperatures.

\textbf{UTKFace (Age Estimation).} The UTKFace dataset comprises facial images annotated with the exact age of each individual. We retain the raw integer ages without grouping into broader bins, thereby treating each age from 0 to 100+ as a distinct ordinal class. The prediction target becomes the exact age label with a precision of \textbf{1 year}. This setup reflects a fine-grained ordinal prediction problem where adjacent age classes (e.g., 31 vs. 32) are semantically closer than distant ones (e.g., 20 vs. 60), preserving the ordered structure of the output space.

\textbf{Avocado Price Dataset.} In this dataset, the target variable is the continuous average price of avocados sold across various U.S. regions and time periods. We convert the continuous price into discrete classes by rounding to the nearest \textbf{0.01 dollar}, resulting in a classification task over ordered price bins. Each class represents a price interval of 1 cent, maintaining the ordinal progression of market price. This formulation enables ordinal-aware models to reason about proximity between predicted and true price levels, improving interpretability and alignment with real-world decision boundaries.

\textbf{IMDB-WIKI (Age Estimation).}
The IMDB-WIKI dataset contains over 500{,}000 facial images annotated with the exact chronological age of each individual at the time the photo was taken. We directly use the raw integer ages without merging into coarse groups, thereby treating each age value (0--100+) as a unique ordinal class. This yields a fine-grained ordinal prediction task with a resolution of 1 year, where adjacent age classes (e.g., 27 vs.\ 28) carry stronger semantic similarity than distant ones (e.g., 15 vs.\ 60). Owing to its large scale, diverse visual conditions, and highly non-uniform age distribution, IMDB-WIKI presents a challenging benchmark for evaluating methods that aim to produce calibrated and contiguous ordinal prediction sets.

\subsection{Conformal Prediction Setup}

We adopt a standard split-conformal prediction setup across all datasets. For each dataset, we precompute 10 random permutations and perform 10 independent trials. In each trial:

\begin{itemize}
    \item The dataset is randomly permuted using a fixed seed and loaded from pre-saved permutation files.
    \item The dataset is split into two equal halves: the first half is used for calibration, and the second half for evaluation.
    \item Model prediction scores and corresponding ground-truth labels are loaded from \texttt{scores.npy} and \texttt{labels.npy}, respectively.
    \item We apply a standard validity check to remove invalid or missing samples before splitting.
\end{itemize}

The predictions are based on softmax probability vectors over the discretized ordinal labels. The first and third experiments are repeated with a fixed miscoverage level $\alpha = 0.1$, and summary statistics (coverage and average prediction set size) are reported as mean and standard deviation across trials.

\subsection{Additional Experiment Result for Hyper-parameter $\lambda$}

To further evaluate the generalizability of our method, we conduct additional experiments on the UTKFace dataset. This dataset involves age estimation, which naturally exhibits an ordinal structure. We assess the effect of the regularization parameter $\lambda$ in our proposed min-RCPS method, examining both the coverage and prediction set size across a range of $\lambda$ values. The results demonstrate consistent trends and further support the effectiveness of regularization in enhancing predictive efficiency without sacrificing coverage.

\begin{figure*}[t]
    \centering
    \includegraphics[width=0.45\linewidth]{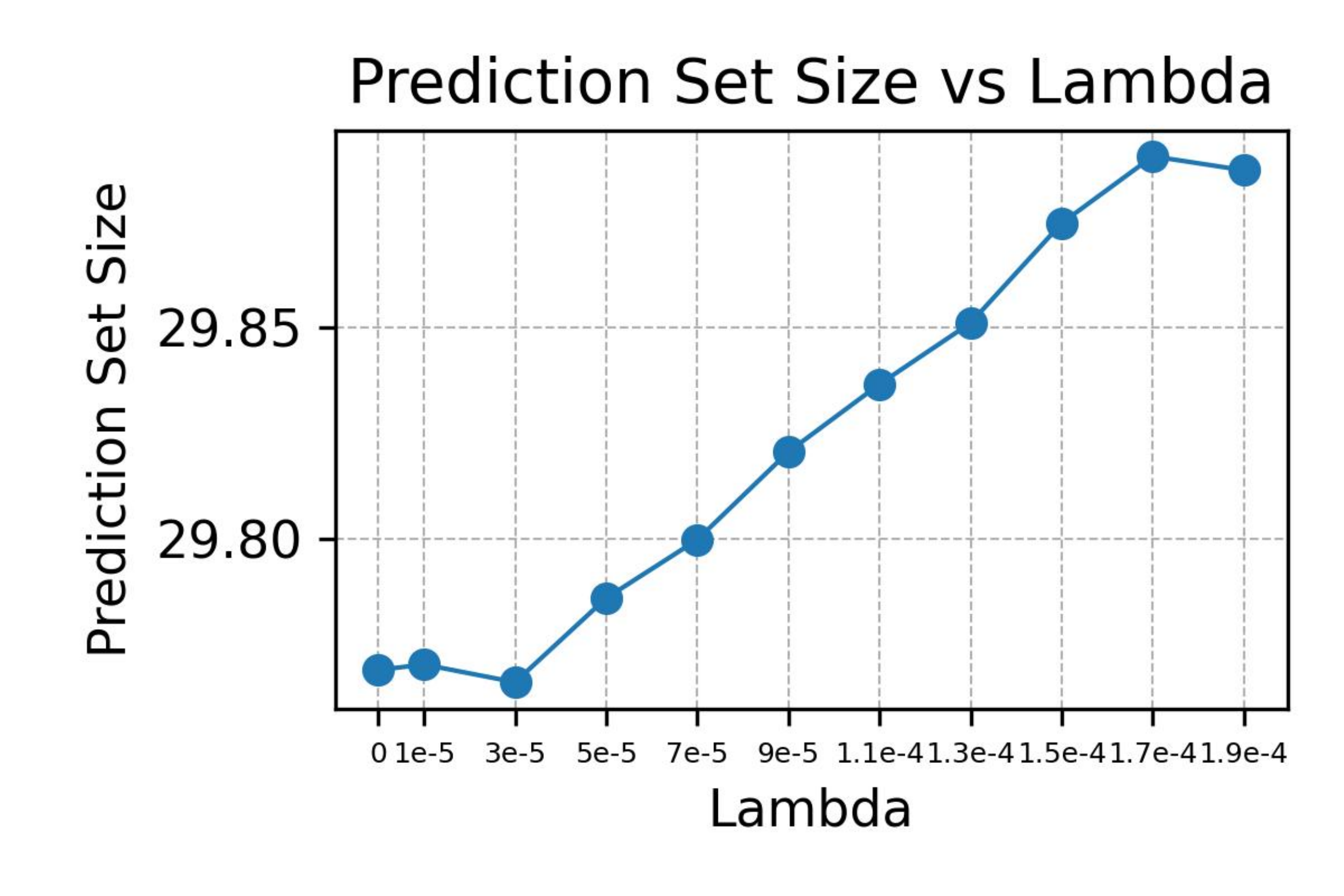}
    \includegraphics[width=0.45\linewidth]{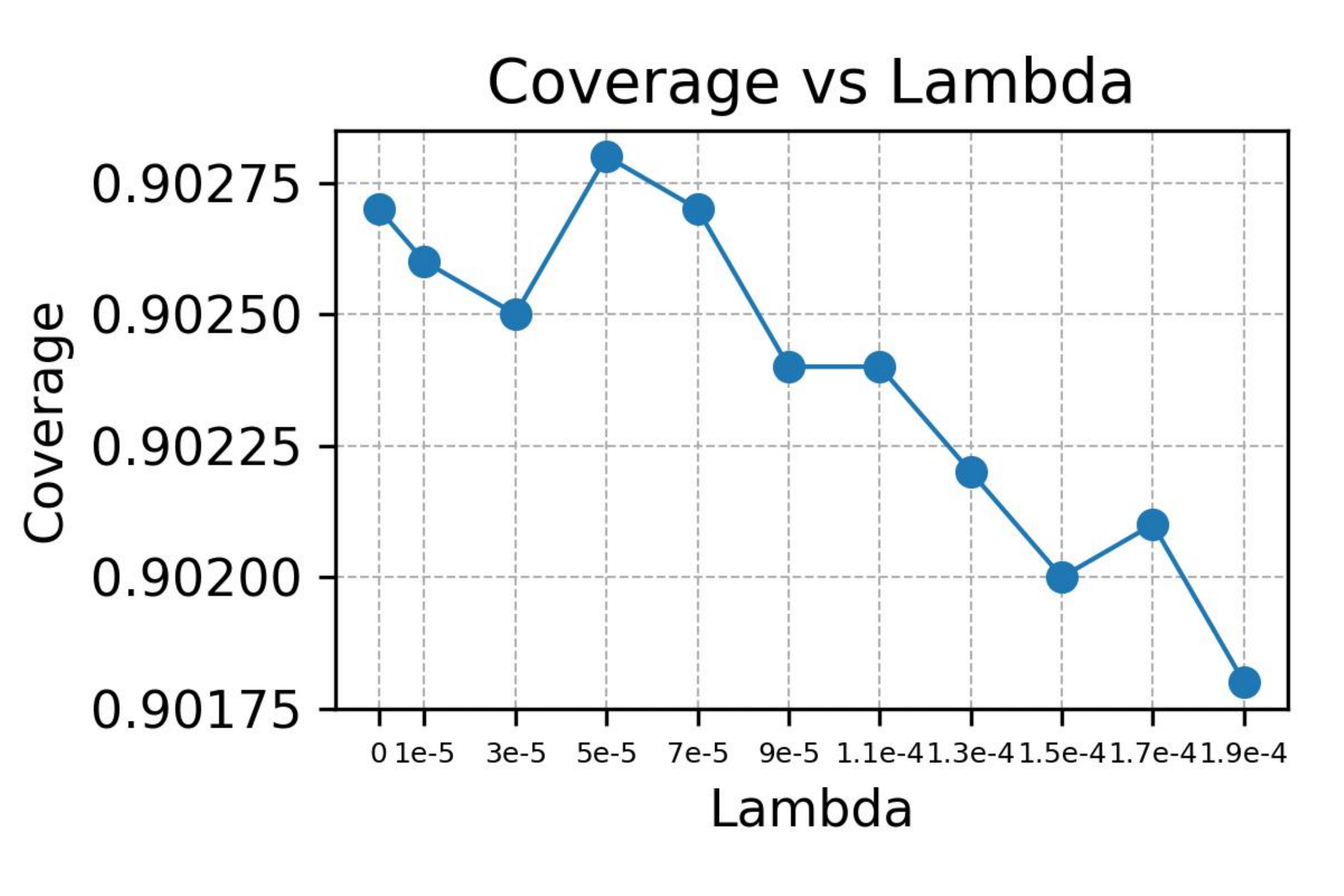}
    \caption{Impact of the regularization hyper-parameter $\lambda$ in min-RCPS on coverage and prediction set size on the UTKFace dataset.
    When the coverage is guaranteed, the prediction set size initially increases as $\lambda$ increases from $0$ to $1e-5$, then decreases as $\lambda$ increases from $1e-5$ to $3e-5$
    where min-RCPS outperforms min-CPS by 0.05\%{$\downarrow$} in terms of prediction set size.
    We find that setting $\lambda$ to a relatively small value is sufficient to improve the predictive efficiency, and this happens on other datasets.
    }
    \label{fig:appendix:lambda_curves}
\end{figure*}